\documentclass{article}

    \usepackage[preprint, nonatbib]{neurips_2021}

\usepackage[utf8]{inputenc} 
\usepackage[T1]{fontenc}    
\usepackage{hyperref}       
\usepackage{url}            
\usepackage{booktabs}       
\usepackage{amsfonts}       
\usepackage{nicefrac}       
\usepackage{microtype}      
\usepackage{wrapfig}
\usepackage{amsmath}
\usepackage{subfig}
\usepackage{graphicx}
\usepackage{amsmath,amssymb}
\usepackage[utf8]{inputenc} 
\usepackage[T1]{fontenc}    
\usepackage{tikz}
\usepackage{blindtext}
\usepackage{enumitem}
\setlist[itemize]{align=parleft,left=0pt..1em}
\usepackage{url}
\usepackage{xcolor}
\usepackage{float}
\usepackage{multirow}
\usepackage{caption}
\usepackage{comment}
\usepackage[titlenumbered,ruled,algo2e]{algorithm2e}
\usepackage{array}
\usepackage{wrapfig}
\usepackage{fdsymbol}

\title{RHNAS: Realizable Hardware and Neural Architecture Search}

\author{Yash Akhauri*\\
Intel Labs, India\\
\And
Adithya Niranjan* \\
Intel Labs, India \\ 
\And
J. Pablo Mu\~{n}oz \\
Intel Labs, USA\\ 
\And
Suvadeep Banerjee \\
Intel Labs, USA\\ 
\And
Abhijit Davare \\
Intel Labs, USA\\ 
\And
Pasquale Cocchini \\
Intel Labs, USA\\ 
\And
Anton A. Sorokin \\
Intel Labs, USA\\ 
\And
Ravi Iyer \\
Intel Labs, USA\\ 
\And
Nilesh Jain\\
Intel Labs, USA\\ }
\begin{document}

\newcommand\TODO[1]{\textcolor{red}{#1}}
\newcommand{\OurScheme}{RHNAS}
\maketitle

\begin{abstract}

The rapidly evolving field of Artificial Intelligence necessitates automated approaches to co-design neural network architecture and neural accelerators to maximize system efficiency and address productivity challenges. To enable joint optimization of this vast space, there has been growing interest in differentiable NN-HW co-design. Fully differentiable co-design has reduced the resource requirements for discovering optimized NN-HW configurations, but fail to adapt to general hardware accelerator search spaces. This is due to the existence of non-synthesizable (invalid) designs in the search space of many hardware accelerators. \textbf{To enable efficient and realizable co-design of configurable hardware accelerators with arbitrary neural network search spaces, we introduce RHNAS. RHNAS is a method that combines reinforcement learning for hardware optimization with differentiable neural architecture search. RHNAS discovers realizable NN-HW designs with  $\mathbf{1.84\times}$ lower latency and $\mathbf{1.86\times}$ lower energy-delay product (EDP) on ImageNet and $\mathbf{2.81\times}$ lower latency and $\mathbf{3.30\times}$ lower EDP on CIFAR-10} over the default hardware accelerator design.

\end{abstract}  

\section{Introduction}
\label{sec:intro}

The success of deep neural networks (DNNs) in fields such as computer vision, recommender systems, and virtual reality is driving an exponential increase in deployment scenarios for deep learning (DL) systems. This success has also been driven by larger, compute-intensive models. These large models have higher latency and energy costs - one of the primary hurdles in developing effective artificial intelligence (AI) systems. To meet the increasing engineering and deployment requirements needed to optimize AI systems, there has been a shift to utilizing automated neural architecture search techniques~\cite{liu2019darts, Hu2020DSNASDN, cai2019proxylessnas}, as well as automated hardware optimization~\cite{Kao2020GAMMAAT, Kao2020ConfuciuXAH}. 
While HW-Aware ML on fixed hardware does improve efficiency,  this is often sub-optimal for configurable accelerators. Such accelerators present significant space for improvement through optimization on components such as number of processing elements, buffer sizes, and data-types, among others. Fixing the NN architecture and optimizing the HW design also leaves performance on the table - different accelerators have different modes of operations and optimizations. Therefore, an optimal layer choice for a given accelerator may not be the best choice for another. To model these inter-dependencies, there has been growing interest in evolutionary algorithm-based ~\cite{Lin2019NeuralHardwareAS}, reinforcement learning-based methods~\cite{Abdelfattah2020BestOB,jiang2019accuracy,yang2020coexploration}, as well as fully differentiable methods~\cite{Zhang2020DNADN, Choi2020DANCEDA} to co-design both hardware accelerators and neural network architectures. 

 Fully differentiable approaches often introduce HW parametrization to represent HW design choices. While some accelerators are fully parametrizable, several of them are \emph{sparsely valid}. Sparse validity can have many reasons, be it Instruction Set Architecture (ISA) limitations, arbitrary constraints in synthesizable designs or NN-HW interactions. In the context of developing HW accelerators and compiler stacks, such issues with validity can arise quite often during early stages of development. Reducing this sparsity can take considerable verification and development efforts, delaying product and innovation life-cycles. We therefore feel that the assumption of fully parametrizable/valid spaces design spaces to be fairly restrictive and hope to address some of these issues with our work. Optimizing on sparsely valid spaces is a strictly harder problem than having non-differentiable hardware costs, as no hardware cost exists for non-realizable designs. We discuss the difficulty of co-design in sparsely valid spaces in more detail in Section \ref{sec:vta_intro}. For the remainder of this paper, we use \emph{non-realizable} and \emph{invalid} hardware designs to refer to choices that are not synthesizable. \\ \\
 In this paper, we look into finding a \emph{realizable} solution in two aspects. First, finding optimal yet synthesizable designs in the hardware accelerator space. Second, proposing a way of co-discovering NN-HW designs with reasonable run-time and compute requirements by leveraging differentiable neural architecture search, something that is becoming increasingly important considering the carbon impact of AI~\cite{Dhar2020}. For example, BestOB~\cite{Abdelfattah2020BestOB} utilizes a RL controller for co-design and requires 1000 GPU hours to co-design NN-HW on the CIFAR-100 dataset.

 We propose a method that integrates a reinforcement learning-based hardware optimizer with differentiable neural architecture search. This method allows us to efficiently explore a prohibitively large NN-HW joint space with non-differentiable hardware costs and sparsely valid hardware designs. \textbf{ With these improvements, we co-design NN-HW on the ImageNet dataset in 186.7 GPU hours and 4 GPU hours on the CIFAR-10 dataset.} In this paper, we discuss the following contributions:
\begin{enumerate}
    \item We introduce \OurScheme{}, an NN-HW co-design method that integrates RL-based hardware optimizers with differentiable NAS. Our method successfully overcomes the challenges associated with sparse validity - a failure point for existing differentiable co-design works. It does so with minimal overhead to a standard DNAS solution.
    \item We benchmark our RL-based hardware optimizer and use Bayesian hyperparameter optimization to identify the best hyper-parameters for a fair study of a range of standard RL algorithms (PPO, A2C, DQN)~\cite{schulman2017proximal, mnih2016asynchronous, mnih2013playing}.
    \item We empirically verify that using fully differentiable approaches to co-design does not aid in discovering valid hardware accelerator designs. We benchmark our RL-based hardware optimizer against a multi-layer perceptron (MLP) based hardware generator~\cite{Choi2020DANCEDA}. Further, we study the convergence of differentiable algorithms which learn optimal hardware parametrizations~\cite{Zhang2020DNADN} on our HW accelerator search space.
\end{enumerate}

\section{Related Work}
\label{sec:to_related_work}

\textbf{Hardware-aware NAS -} Early efforts towards the discovery of NN architectures primarily focused on optimizing accuracy using reinforcement learning (RL) and evolutionary algorithms (EA) ~\cite{howard2019searching, tan2019mnasnet, tan2020efficientnet}. These algorithms generally require considerable computational resources and need to evaluate and train many candidate architectures to completion. Differentiable NAS strategies were introduced to reduce the cost of discovering efficient neural networks~\cite{Liu2019DARTSDA, Hu2020DSNASDN}. Most of the early focus was on maximizing accuracy, which leads to very large neural networks with high latency and energy requirements. To address these demanding requirements, many works ~\cite{cai2019proxylessnas, Wu2019FBNetHE, xie2020snas, Xu2020LatencyAwareDN, Zhang2020FastHN} model the latency, floating-point operations, or other performance-related metrics as regularizers to guide NAS. 

\textbf{Hardware design optimization -} DNN accelerators~\cite{han2016eie, Umuroglu_2017, umuroglu2020logicnets, chen2019eyeriss, 10.1145/3093337.3037702, 10.1145/2654822.2541967} have been developed with the prospect of potential gains over general-purpose hardware. Eyeriss is an energy-efficient CNN accelerator~\cite{Chen2019EyerissVA} that delivered 10 times the energy efficiency over mobile GPUs. Such improvements lead to further research in novel microarchitectures and data flow styles. To get the best performance on a range of workloads, many accelerators are being designed for configurability based on the selected workload. This results in many possible hardware resource assignments dictated by the NN topology to be deployed. Several approaches have been proposed to identify optimal resource assignments, e.g., ConfuciuX~\cite{Kao2020ConfuciuXAH}, GAMMA~\cite{Kao2020GAMMAAT}, MARVEL~\cite{chatarasi2020marvel} utilize RL and GA (Genetic Algorithms) to decide the hardware resource assignment for a given workload. 

\textbf{Hardware-Neural Network Co-design -} While NAS and HW design optimization give good results when applied standalone, the significant inter-dependencies between NN and HW components suggest an opportunity for improvement by jointly modeling NN topologies and HW designs. DANCE~\cite{Choi2020DANCEDA} introduces an MLP based HW generator along with a differentiable performance estimator to guide the joint differentiable search. DNA~\cite{Zhang2020DNADN} converts the hardware and neural network design space into a categorical probability distribution and uses an MLP HW performance predictor for the co-design of hardware and neural network. We find that these fully differentiable solutions cannot adapt to spaces that has many invalid hardware design choices. Some works propose an RL controller for co-design~\cite{Abdelfattah2020BestOB, Lu2019OnNA}. However, these works suffer from high search cost problems~\cite{Choi2020DANCEDA}. 
\section{Configurable Hardware Accelerators and Sparsely Valid Design Spaces}
\label{sec:vta_intro}

We use the Versatile Tensor Accelerator (VTA) as our configurable hardware accelerator. VTA is a programmable deep learning accelerator~\cite{moreau2019hardwaresoftware} with parametrizable architecture paired with Apache \textbf{\textbf{}}TVM~\cite{chen2018tvm}, a deep learning compiler stack. This integrated end-to-end deep learning stack provides a flexible platform to implement our NN-HW co-design framework. The VTA HW is designed in \textit{Constructing Hardware in a Scala Embedded Language} (CHISEL)~\cite{bachrach2012chisel} that facilitates full parametrization where underlying memory elements can be programmatically customized for the HW backend (ASIC/FPGA), but the hardware designer also retains the ability to reuse existing hardware components. The CHISEL flow generates HW modules expressed in register-transfer language (RTL), which can be used for downstream ASIC/FPGA synthesis. The TVM framework utilizes a HW simulation guided schedule auto-tuning platform known as autoTVM~\cite{chen2018learning} for computing optimal data access patterns in adaptation to changes in the workload and HW configurations. Details on how we calculate latency and energy-delay product (EDP) are provided in the appendix. 

For this work, we have modified the TVM compiler to utilize an analytical bandwidth minimization technique to compute the optimal tiling scheme of NN operator schedules rapidly. In addition, we have extended the TVM-VTA stack to support depth-wise convolutions to allow simulation of our NN search space. Table \ref{tab:hw_space} shows the HW architecture parameters we have considered and their possible values (in $log_{2}$ scale). The product of $block\_in$ and $block\_out$ decides the size of the GEMM core of the accelerator, the input (Inp), weight (Wgt), and accumulator (Acc) buffers are on-chip SRAMs, $\mu$Op width defines the size of the RISC micro-ops, and the $\mu$Op buffer defines the size of the micro-op cache. 

\begin{wraptable}{r}{0.3\linewidth}
  \centering
  \caption{VTA parameter ranges ($log_{2}$).}
  \label{tab:hw_space}
{
        \begin{tabular}{cc}
        \toprule
        \textbf{HW Feature} & \textbf{Range} \\
        \midrule
        block\_in & [3, 6] \\
        block\_out & [3, 6] \\
        $\mu$Op Width (bits) & [5, 6] \\
        $\mu$Op Buffer (kB) &  [5, 6]\\
        Inp Buffer (kB) & [13, 20]\\
        Wgt Buffer (kB) & [13, 20]\\
        Acc Buffer (kB) & [13, 20]\\
        \bottomrule
        \end{tabular}
    }
\end{wraptable}

While independent consideration of each architectural parameter in Table \ref{tab:hw_space} leads to a large theoretical design space (32768 possible configurations), many parameter combinations are invalid. This is a common problem on hardware accelerators. On VTA, it is due to ISA restrictions. For example, large compute cores and local memory buffers can improve accelerator performance, but the number of bits required to address such cores/buffers may exceed the instruction width, thus limiting the scope of joint architectural parameter exploration. Another issue involves hardware resource limitations. For example, FPGA targets may lack the SRAM required, or ASIC synthesis may not reach the specified frequency. There can also exist arbitrary points in the parameter space which are not synthesizable due to routing congestion of larger designs. While such designs are within the parameter space, it is not possible to realize them. As configurable accelerators may have invalid parameter choices, we implement a validity checker function that analytically detects whether a hardware design is valid.

\section{The \OurScheme{} Flow}
\label{sec:flow}

\subsection{Proposed Co-Exploration}
\label{subsec:proposal}

\begin{figure*}[!htbp]
\begin{center}
    \centerline{\includegraphics[width=\columnwidth]{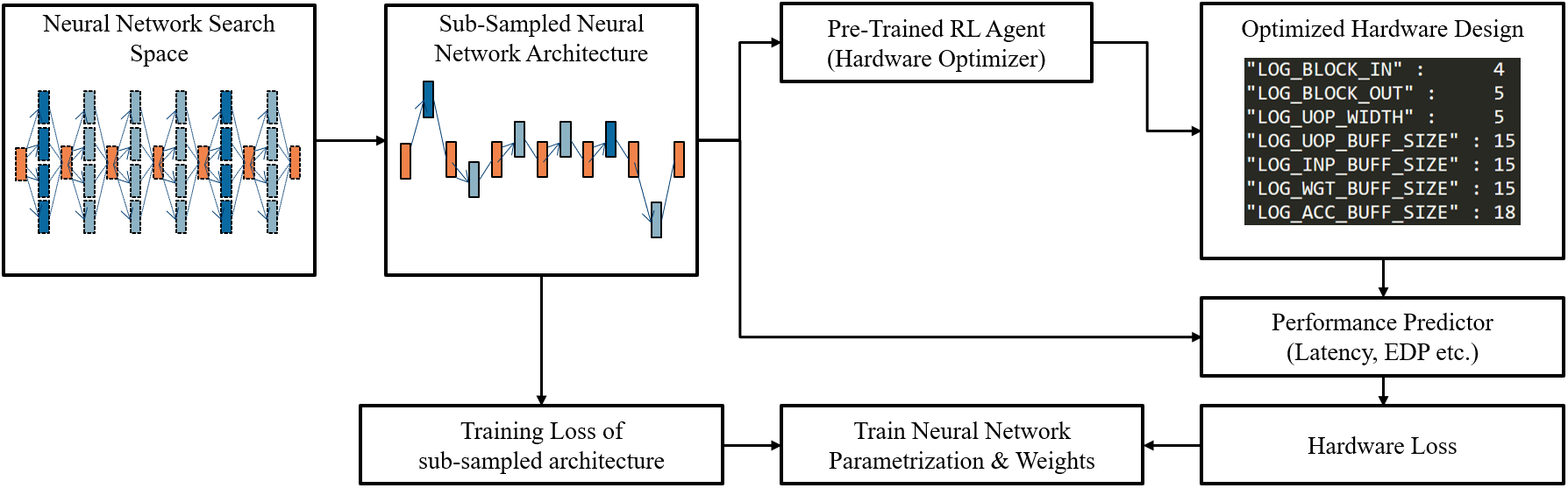}}
\end{center}
	\caption{Overall \OurScheme{} flow diagram}
    \label{fig: overall_flow}
  \vspace{-0.5cm}
\end{figure*}

We introduce an RL HW optimizer whose training is disentangled from differentiable NAS to address the dual problem of sparsely valid hardware design space and high NN architecture search cost associated with RL and EA-based NAS algorithms. Figure \ref{fig: overall_flow} shows an overview of our \OurScheme{} flow. Our differentiable NAS algorithm samples neural network architectures from a super-net. This sampling is done from a super-net using a categorical architectural distribution. We use a pre-trained RL HW optimizer to generate an optimized accelerator configuration for the sub-sampled NN architecture. For every NN architecture sampled during the differentiable neural architecture search process, the RL HW optimizer receives the subsampled NN architecture along with a default HW accelerator configuration template and outputs an optimized HW accelerator configuration. A performance predictor is then used to get the performance of the optimized NN-HW design. This performance is used as a regularizer in the differentiable NAS minimization objective.

\begin{wraptable}{r}{0.45\linewidth}
\caption{\OurScheme{}'s Optimization}
\label{tab:nascha_opt_eqns}
    \begin{tabular}{c}
        \toprule
    $\min \limits_{\alpha, \theta} \,\, L_{train}( Z_{\alpha, \theta})+\lambda L_{hw}(\sigma(\alpha), H^{opt})$ \label{eq:update_alpha}\\
    $s.t. \quad L_{hw} = \text{PerfPredictor}(\sigma(\alpha), H^{opt})$ \label{eq:hw_optimizer_rl}\\
    $s.t. \quad H^{opt} = \text{HWOpt}(\sigma(\alpha), H_{0})$ \label{eq:hw_optimizer_op}\\
        \bottomrule
    \end{tabular}
\end{wraptable}

\OurScheme{}'s optimization is formulated as shown in Table \ref{tab:nascha_opt_eqns}. $\sigma$ is the Softmax function. $\alpha$ is the continuous variable parametrization of the NN architecture. $\sigma(\alpha)$ is the categorical probability distribution of the NN supernet. $Z_{\alpha, \theta}$ is the one-hot representation of the subnetwork sampled using the $\sigma(\alpha)$ and $\theta$ weights. HWOpt is the RL HW optimizer. $H_{0}$ is the template HW accelerator configuration that the HWOpt is initially provided with. If HWOpt generates an invalid accelerator configuration, it simply outputs the template HW accelerator configuration $H_{0}$. $L_{hw}$ is the HW cost loss determined by the NN-HW Performance Predictor (e.g., Clock Cycles, Energy-Delay-Product) and is used to guide the neural architecture search.

As indicated in  Table \ref{tab:nascha_opt_eqns}, we use the NN architecture categorical distribution $\sigma(\alpha)$ instead of the sub-sampled one-hot representation $Z_{\alpha, \theta}$ for the HW loss. In our tests, using the one-hot architectural choice $Z_{\alpha, \theta}$ in Table \ref{tab:nascha_opt_eqns} leads to noisy training, whereas $\sigma(\alpha)$ leads to smoother training as it is a continuous probability distribution. 
\subsection{HW Performance Predictors}
\label{subsec:diffmlp}
  
For any given accelerator configuration and neural network architecture, the simulator takes significant time to output a performance metric, as seen in Table \ref{tab:sim_outputs}. For example, we need 2 minutes to get a single cycle-accurate RTL simulation of an end-to-end ResNet50v2model. The large query time makes computing HW metrics on the fly impractical. Metrics such as FLOPs are not capable of capturing the intricacies of the NN-HW design space, we find that FLOPS are weakly correlated with cycle count. To quantify this - the Pearson's correlation coefficient between cycle count and FLOPs for 55420 randomly sampled NN-HW design choices was 0.15 (weakly positive correlation).

We use an MLP to avoid large query times and capture the relations between NN-HW designs and performance. The MLP takes as input the joint NN-HW embedding and predicts the Latency/EDP. As seen in Table \ref{tab:nascha_opt_eqns}, the NN embedding the MLP takes is a probability distribution ($\sigma(\alpha)$) instead of a one-hot encoding. We include a study in the appendix to validate that the performance predictor is able to accurately interpolate between different discrete NN architectural choices.
\subsection{Reinforcement Learning for HW Optimization}

\label{subsec:hwopt}
We use an RL agent to optimize over our sparsely valid hardware design space. The agent receives a representation of the neural network ($N$) and an initial HW accelerator template ($H_{0}$) of our accelerator. The agent modifies the accelerator configuration to improve the NN-HW design latency/EDP. In order to make the agent minimize the latency/EDP, we provide the negative of the predicted latency/EDP as the reward. In a sparsely valid design space, optimizing only latency/EDP can lead to invalid accelerator designs. To address this challenge, we penalize the agent with $C_{inv}$ for providing invalid accelerator design parameters. We study two intuitive settings based on this formulation:

\textbf{Composite setting}:  In this formulation, we use a concatenated representation of the NN and HW $(N, H_{t})$ as the environment state. The agent takes a discrete action that specifies the whole HW configuration ($H_{t}$) every time step and runs for a fixed number of timesteps $t_{max}$ every episode. As the composite agent provides a new hardware configuration in each time step, it may lose incentive to optimize a suboptimal accelerator configuration due to penalty for invalid states. To encourage continuous improvement, we introduce a penalty $B$ imposed when the accelerator configuration the composite agent provides is worse than its previous accelerator configuration ($P(N, H_{t}) > P(N, H_{t-1}$). \\
\textbf{Sequential setting}: In this approach, the RL agent optimizes only one HW accelerator parameter in a time step, making $t_{max}$ passes in an episode. Here, we modify the state to include a one-hot vector $S_{t}$ to provide the agent context about which parameter to optimize. At each time step, the agent takes a discrete action choice ($H^{p}_{t}$ for the $p^{th}$ step). After the agent has generated an action for every HW parameter (we have 7 parameters) $t_{max}$ times, we get the final HW configuration. In this setting, we found that providing a reward after generating a full template to be more stable than rewards at every time-step.\\
We discuss implementation details of the agent and the state representation further in Section 5.2 and the appendix.
\begin{table}[hbt!]
\vspace{-4mm}
\caption{RL formulation. $O_{t}$, $A_{t}$, $R_{t}$ are state, action and reward respectively}
\label{tab:rl_formulation}
    \begin{tabular}{ll}
        \toprule
Composite setting & Sequential setting\\
        \midrule
    $O_{t} = (N, H_{t})$; \;\;\;\;\;    & $O_{t} = (N, H_{t}, S_{t})$; \;\;\;\;   \\$A_{t} = (H_{t})$
     &  $A_{t} = (H^{p}_{t}) \quad s.t. \quad    H^{p}_{t} \in [0,7]$  \\  $R_{t} =
    \begin{cases}
 &  -P(N, H_{t}) - C_{inv}(H_{t}) \text{, if  }  t =  t_{max} \\ 
 & -B \text{, if  }   P(N, H_{t}) > P(N, H_{t-1})\\
 & 0 \text{,  otherwise} \\
\end{cases}$ & $R_{t} =
    \begin{cases}
 &  -P(N, H_{t}) - C_{inv}(H_{t}) \text{, if  }  S_{7}= 1 \\ 
 & 0 \text{, otherwise}
\end{cases}$\\
\bottomrule
    \end{tabular}
\end{table}
\section{Evaluation of RHNAS Components}
\label{sec:eval}

\subsection{Performance Predictor Network}
\label{subsec:predictor_eval}
\begin{wraptable}{r}{0.4\linewidth}
  \vspace{-0.4cm}
  \caption{Testing the predictor network performance by measuring the Kendall Tau rank correlation coefficient. ($\tau$)}
  \label{tab:sim_outputs}%
  \centering
{
        \begin{tabular}{cc}
        \toprule
        Predictor Network& $\tau$ metric \\
        \midrule
        EDP                  & 0.982 \\
        Cycle Count          & 0.982 \\
        GEMM Cycle Count    & 0.984 \\
        $\mu$Op load (bytes) & 0.975 \\
        Weight load (bytes) & 0.985 \\
        Input load (bytes)   & 0.983 \\
        Accumulator load (bytes) & 0.984 \\
        \bottomrule
        \end{tabular}%
    }
\end{wraptable}
Our performance predictor network is a fully connected neural network with 3 layers with 512 neurons in each hidden layer and intermediate ReLU activation functions. Table \ref{tab:sim_outputs} shows the clock cycles, EDP and other simulation metrics. We used the VTA simulator to randomly sample 53600 NN-HW pairs to generate our dataset. We train our networks with the L1 loss between predicted and actual performance. Our design space changes the buffer and GEMM core sizes by powers of two. There can be a large difference in the clock cycles for different NN-accelerator designs. We found that the \emph{Robust Scaler} normalization from sklearn~\cite{scikit-learn} works well on our data. Since we use the performance predictor output to guide our search, it is important that the ranking of different NN-HW architecture pairs are preserved. To verify this, we use the Kendall rank correlation coefficient ($\tau$) ~\cite{10.2307/2332226} - a $\tau$ of 1 indicates that the rank ordering is perfectly preserved. Since we find that $\tau > 0.97$, we conclude that our predictors can be used in the \OurScheme{}.
\subsection{Hardware Optimizers}
\label{subsec:hwopt_eval}

For the RL HW optimizer, we experimented with the Proximal Policy Optimization (PPO) \cite{schulman2017proximal}, Advantage Actor-Critic (A2C) \cite{mnih2016asynchronous}, and Deep Q-Network (DQN) \cite{mnih2013playing} algorithms. 
We implemented the composite and sequential action spaces and environments with the OpenAI gym environment and use implementations of PPO, A2C and DQN algorithms from the stable-baselines3 library \cite{stable-baselines3}.

\begin{wraptable}{r}{0.5\linewidth}
  \centering
  \caption{Optimality of HW generated by RL HW generators.}
  \label{tab:rl_agent}%
{
        \begin{tabular}{ccc}
        \toprule
        \textbf{Cycle Count}  & \multicolumn{1}{c}{\textbf{Composite}} & \multicolumn{1}{c}{\textbf{Sequential}} \\
        \midrule
        PPO & \textbf{99.82}\% & 99.79\% \\
        A2C & 99.82\% & 56.51\% \\
        DQN & -  & 98.10\% \\
        \toprule 
        \textbf{EDP}  & \multicolumn{1}{c}{\textbf{Composite}} & \multicolumn{1}{c}{\textbf{Sequential}} \\
        \midrule
        PPO & \textbf{98.61}\% & 97.56\% \\
        A2C & 97.53\% & 44.05\%  \\
        DQN & - & 95.21\%  \\
        \bottomrule
        \end{tabular}%
    }
\end{wraptable}

For the RL environment, we treat the validity penalty ($C_{inv}$), maximum timesteps ($t_{max}$), and the intermediate penalty ($B$), as environment hyper-parameters. We found that tuning these parameters carefully was key to finding agents that generated both optimal and synthesizable hardware configurations .For example, for penalty values that were too low ($C_{inv}$ < 1.5), the RL agent would generate very large accelerator configurations, which do not fit on our target hardware. On the other hand, larger penalties meant the agent would quickly stop at a valid sub-optimal HW configuration. Using HPO (details in appendix), we were able to find agents that give near-optimal synthesisable configurations. From Table \ref{tab:rl_agent}, we observe that PPO~\cite{schulman2017proximal} with the composite setting gives close to optimal configuration ($\sim99\%$ optimal) for both cycle count and energy delay product (EDP) while generating no invalid configurations 

\subsubsection{Ablation Study on Hardware Optimizers}
\label{subsubsec:hwopt_ablation}
    A recent work, DANCE~\cite{Choi2020DANCEDA} introduced an MLP based hardware generator. The data-set is generated by sampling random NNs from the search space and using the output of a hardware generation toolkit as the target HW. This toolkit uses exact algorithms such as exhaustive search or branch-and-bound algorithms to output optimized HW configurations for an input NN. We reference this method as \textit{Exhaustive-HWGEN} in Table \ref{tab:mlp_agent}. In our tests, Exhaustive-HWGEN is a 3 layer MLP with 512 neurons and is trained by taking a categorical cross entropy loss with respect to the optimal HW design for a given NN architecture. The optimal HW design is found by exhaustive search. While this method slightly exceeds the optimality of hardware generated by our reinforcement learning agents, it has a significantly larger training time for a general accelerator space because of the exhaustive search. Exhaustive search is not practically feasible as accelerator designs get increasingly complex. If we look at most accelerators, the search space is of the order of $10^{30}$ to $10^{70}$~\cite{Zhang2020DNADN, Kao2020ConfuciuXAH} choices, generating the data-set for Exhaustive-HWGEN would be completely infeasible.

\begin{wraptable}{r}{0.5\linewidth}
  \caption{Optimality of HW generated by MLP HW generators.} 
  \label{tab:mlp_agent}%
  \centering
  {
        \begin{tabular}{ccc}
        \toprule
        \textbf{Cycle Count} & \textbf{Optimality} &\textbf{\% invalid} \\
        \midrule
        Exhaustive-HWGEN & \multicolumn{1}{c}{99.9\%} & 0\% \\
        Perf-HWGEN & \multicolumn{1}{c}{N/A} & 100\% \\
        \toprule
        \textbf{EDP} & \textbf{Optimality} & \textbf{\% invalid} \\
        \midrule
        Exhaustive-HWGEN & \multicolumn{1}{c}{99.4\%} & 0.8\% \\
        Perf-HWGEN & \multicolumn{1}{c}{N/A} & 100\% \\
        \bottomrule
        \end{tabular}%
    }
\end{wraptable}
As an alternative to exhaustive search, we test \emph{Perf-HWGEN}, which is a hardware generator akin to Exhaustive-HWGEN but with a training procedure that does not require exhaustive enumeration. We utilize the performance predictor MLP from Section \ref{subsec:predictor_eval} in our loss formulation for the Perf-HWGEN hardware generator. Since we are not using exhaustive enumeration as a search strategy, we also have to penalize invalid configurations given by the Perf-HWGEN. For this purpose, we use \emph{ValidNet MLP}. This MLP indicates whether an input HW accelerator configuration is valid. Details on ValidNet MLP architecture and training is provided in the appendix. 

The training loss for Perf-HWGEN is formulated as,

\begingroup
\allowdisplaybreaks
  \vspace{-0.5cm}
\begin{align} 
    \begin{split}
    & \min \limits_{\beta} \,\, L_{hw}(Z, \text{Perf-HWGEN}_{\beta}(Z)) + \lambda L_{valid}(\text{Perf-HWGEN}_{\beta}(Z)) \label{eq:hwgenmlp_loss} 
    \end{split}
\end{align}
\endgroup
As indicated in Equation \ref{eq:hwgenmlp_loss}, the training objective of Perf-HWGEN is to jointly minimize the hardware cost ($L_{hw}$) and validity loss ($L_{valid}$). $Z$ is a randomly sampled NN architecture and $\beta$ are the parameters of Perf-HWGEN. If $\lambda = 0$, there is no penalty for invalid configurations. In such cases, the Perf-HWGEN generates invalid configurations by merely maximizing the performance.

Our Perf-HWGEN uses the $L_{valid}$ loss to guide the search for valid configurations. We sweep $\lambda \in [10^{-4}, 10^{2}]$ but do not converge to a Perf-HWGEN that generates valid configuration. We study the behavior of ValidNet in guiding search for valid hardware configurations in Section \ref{subsubsec:fulldiffjointsearch}. It is also important to note that the performance predictor network has only been trained on valid configurations, thus the optimality numbers for Perf-HWGEN are not reliable, and labeled as \emph{Not Applicable (N/A)} in Table \ref{tab:mlp_agent} for invalid configurations.

\begin{table}[!t]
\begin{minipage}{.5\linewidth}  
  \caption{
\OurScheme{} Ablation studies. EDP has been normalized, Lat is Latency.
}
  \label{tab:ablation_table}
    \begin{tabular}{cccc}
    \toprule
    \textbf{Target}  & \textbf{Acc} & \textbf{Lat} & \textbf{EDP} \\
    \textbf{ASIC} & \textbf{ (\%)} & \boldmath{$ms$} &   \\
    \midrule
    DSNAS & 74.4 & 22.13 & 28 \\
    HW Aware NAS & 74.1 & 23.67 & 31 \\
    Sequential Optimization & 74.2 & 13.54 & 17 \\
    \textbf{\OurScheme{} Lat} & \textbf{74.5} & \textbf{12.03} &  \textbf{16} \\
    \textbf{\OurScheme{} EDP} & \textbf{74.5} & \textbf{12.73} & \textbf{15} \\
    \OurScheme{} Lat+FLOPs & 73.4 & 11.77 & 12 \\
    DSHWNAS & 74.2 & \multicolumn{2}{c}{\textit{Invalid Design}} \\
    \bottomrule
    \end{tabular}
\end{minipage}
\hspace{0.05\linewidth}
\begin{minipage}{.45\linewidth}
  \caption{
Evaluating \OurScheme{} with other co-design works.}
  \label{tab:comparision_wsota}
    \begin{tabular}{cccc}
    \toprule
    \textbf{Target}  & \textbf{Latency} & \textbf{Area} & \textbf{Acc} \\
     \textbf{ASIC} & \boldmath{$ms$}& \boldmath{ ($mm^2$)} & \textbf{ (\%)} \\
    \midrule
    \multicolumn{1}{c}{DANCE} & \multicolumn{1}{c}{8.13} & \multicolumn{1}{c}{2.73} & \multicolumn{1}{c}{68.70}  \\
    \multicolumn{1}{c}{DNA 4bit} & \multicolumn{1}{c}{1.25} & \multicolumn{1}{c}{5.5} & \multicolumn{1}{c}{71.7}  \\
    \multicolumn{1}{c}{DNA 16bit} & \multicolumn{1}{c}{2.85} & \multicolumn{1}{c}{2.12} & \multicolumn{1}{c}{72.2}  \\
    \multicolumn{1}{c}{NHAS} & \multicolumn{1}{c}{1.58} & \multicolumn{1}{c}{5.87} & \multicolumn{1}{c}{70.74}  \\
    \multicolumn{1}{c}{EDD} & \multicolumn{1}{c}{11.15} & \multicolumn{1}{c}{-} & \multicolumn{1}{c}{74.7}  \\

    \multicolumn{1}{c}{\OurScheme{}} & \multicolumn{1}{c}{12.03} & \multicolumn{1}{c}{1.21} & \multicolumn{1}{c}{74.5} \\
    \bottomrule
    \end{tabular}
\end{minipage}%
\end{table}

\section{\OurScheme{} Evaluation}
\label{subsec:hwnas_ablation}

\label{subsec:hwnas_eval}

\begin{wraptable}{r}{0.4\linewidth}
  \caption{Comparing against other efficient neural networks.} 
  \label{tab:nasnet_mbv2_rhnas}%
  \centering
  {
        \begin{tabular}{ccc}
    \toprule
    \textbf{Target}  & \textbf{Latency} & \textbf{Acc} \\
     \textbf{ASIC} & \boldmath{$ms$} & \textbf{ (\%)} \\
    \midrule
    \multicolumn{1}{c}{MNASNet-1} & \multicolumn{1}{c}{17.31} & \multicolumn{1}{c}{75.20} \\
    \multicolumn{1}{c}{MobileNetV2} & \multicolumn{1}{c}{12.16} & \multicolumn{1}{c}{71.66} \\
    \multicolumn{1}{c}{\OurScheme{}} & \multicolumn{1}{c}{12.03} & \multicolumn{1}{c}{74.5} \\
    \bottomrule
    \end{tabular}%
}
\end{wraptable}
For our neural architecture search strategy, we employ the super-net and choice of blocks as used in the released implementation of DSNAS~\cite{Hu2020DSNASDN}. Our NN architecture sampling, hyper-parameter settings and parameter update strategy is the same as DSNAS~\cite{Hu2020DSNASDN}. DSNAS introduces a task-specific direct neural architecture search without parameter retraining. DSNAS requires 420 GPU hours to discover networks with accuracy comparable to two-stage (search and train) methods. As we train on an NVIDIA V-100 GPU, we complete the search in 187 GPU hours.

We have conducted several experiments on \OurScheme{} using the ImageNet (ILSVRC2012)~\cite{imagenet_cvpr09} and CIFAR-10~\cite{Krizhevsky2009LearningML} data-sets. We include in our evaluation three ASIC baselines: DANCE~\cite{Choi2020DANCEDA}, DNA~\cite{Zhang2020DNADN} and NHAS~\cite{Lin2019NeuralHardwareAS}. Due to differences in accelerator designs and the neural network backbone, it is not feasible to fairly compare the latency and EDP without further standardization. The HW metrics for different works are largely different based on the capabilities and optimization of the accelerators. We draw the attention of the reader to the fact that DNA cannot discover valid designs on our hardware search space, and is studied in detail by our DSHWNAS test in Table \ref{tab:ablation_table} and Section \ref{subsubsec:fulldiffjointsearch}. The MLP-based hardware generator in DANCE has been studied in Section \ref{subsubsec:hwopt_ablation}. DANCE can work on our hardware design space by exhaustive enumeration, which is not a scalable approach as the complexity of hardware accelerators increase. To address the lack of standardization across accelerators, we deploy MNASNet-1~\cite{pytorch_nets} and MobileNetV2~\cite{pytorch_nets} on the same VTA accelerator as discovered by \OurScheme{} Latency in Table \ref{tab:nasnet_mbv2_rhnas}. We also demonstrate the effectiveness of \OurScheme{}, by implementing HW-Aware NAS and NN-HW sequential optimization methods on our accelerator search space as shown in Table \ref{tab:ablation_table}. All our networks have been quantized to 8 bits. 

\subsection{\OurScheme{} on CIFAR-10}
\label{subsec:nascha_cifar10}

\begin{wrapfigure}{r}{0.5\linewidth}
\centering
\includegraphics[width=.45\columnwidth]{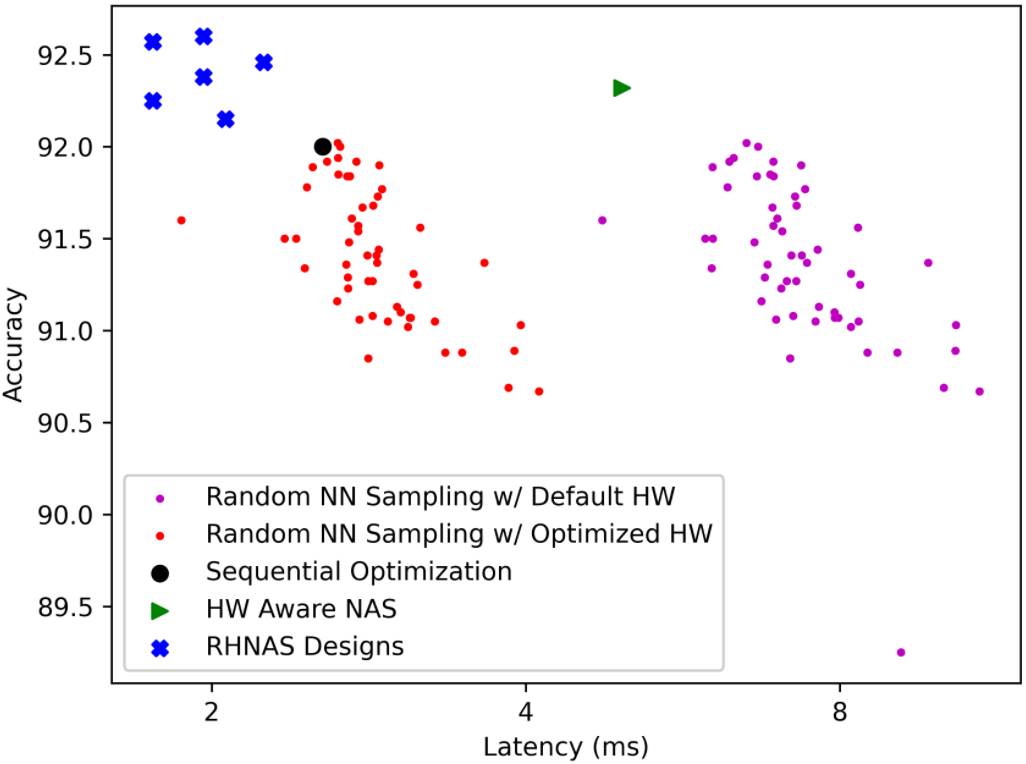}
\caption{\OurScheme{} discovers efficient NN-HW choices that give high accuracy on CIFAR-10.}
\label{fig:nascha_acc_cifar}
\vspace{-5mm}
\end{wrapfigure}

To demonstrate the effectiveness of \OurScheme{} in discovering solutions with low latency and high accuracy, we use a smaller variant of the DSNAS network with 9 layers instead of 20 layers and try to find good HW-NN pairs on the smaller network. There are $4^{9} = 262144$ neural networks in our CIFAR-10 search space. We randomly sample 45 neural networks and train them using the DSNAS~\cite{Hu2020DSNASDN} SPOSretrain240 strategy detailed in the appendix along with more experimental details.   We measure the latency of these neural networks on the default VTA accelerator configuration and are represented by "Random NN Sampling w/ Default HW". We then measure the latency of these neural networks on the optimized HW accelerator design generated by the RL HW Optimizer referenced as "Random NN Sampling w/ Optimized HW". Sequential Optimization refers to a method where we first optimize the neural network architecture for accuracy, and use the RL HW Optimizer to generate the hardware accelerator design for the neural network. We also include HW-Aware NAS, which attempts to find an optimal neural network architecture on the default HW accelerator design. As seen from Figure \ref{fig:nascha_acc_cifar}, \OurScheme{} designs outperform all other methods provided in the figure. 

\subsection{Hardware-Aware NAS}

In Hardware-Aware NAS, we try to minimize cycle count and EDP without a HW optimizer in the loop. 
As described by Equation \ref{eq:update_alpha_nohwopt}, we utilize the default HW configuration and optimize the NN parametrization on the fixed HW. \OurScheme{} finds solutions with $1.97\times$ lower latency and $2.07\times$ lower EDP with a $0.4\%$ accuracy gain on ImageNet and  $2.81\times$ lower latency and $3.30\times$ lower EDP on CIFAR-10.

\begingroup
\allowdisplaybreaks
\begin{align} 
    \begin{split}
    & \min \limits_{\alpha, \theta} \,\, L_{train}( Z_{\alpha, \theta})+\lambda L_{hw}(\sigma(\alpha), H_{0}) \label{eq:update_alpha_nohwopt} 
    \end{split} 
\end{align}
\endgroup
\subsection{Sequential Optimization}
One of the primary points of comparison for a joint search algorithm is to benchmark with sequential optimization algorithms. Since our RL HW optimizer agent is disentangled from the NN architecture search, we can take the resultant network from the NAS algorithm with no simulation/FLOPs loss and generate HW optimized for that topology. Further, we also simulate the original DSNAS algorithm on our HW. As seen from Table \ref{tab:comparision_wsota}, \OurScheme{} Latency is able to find a solution with $1.13\times$ lower latency for $0.1\%$ higher accuracy on the ImageNet dataset and $1.5\times$ lower latency on CIFAR-10 dataset.

\subsection{Fully Differentiable Joint Search}
\label{subsubsec:fulldiffjointsearch}
Several works~\cite{Zhang2020DNADN, Choi2020DANCEDA} introduce a fully differentiable approach to NN-HW co-design. While these works produce excellent results in accelerator design spaces with sparse optimas, we attempt to reproduce a generalized fully differentiable search algorithm as shown in Equation \ref{eq:update_alpha_ablation_fulydiff} and demonstrate its inability to navigate our sparsely valid accelerator design space. Our joint differentiable architecture search is formulated below. We use the same $L_{valid}$ as described in Equation \ref{eq:hwgenmlp_loss}.
\begingroup
\allowdisplaybreaks
\begin{align} 
    \begin{split}
    & \min \limits_{\alpha, \gamma, \theta} \,\, L_{train}( Z_{\alpha, \theta})+\lambda L_{hw}(\sigma(\alpha), \gamma) + \beta L_{valid}(\gamma) \label{eq:update_alpha_ablation_fulydiff}
    \end{split} 
\end{align}
\endgroup

As seen in Table \ref{tab:comparision_wsota}, this DSHWNAS formulation converges with an accuracy of 74.2\%, \begin{wraptable}{r}{0.45\linewidth}
\caption{Gradient Interpolation Study}
\label{eq:hw_parametrization_validnet_update}
    \begin{tabular}{c}
    \toprule
    $\Delta_{j} = \frac{\partial}{\partial \widetilde{\gamma_{j}}}\text{L1Loss}(\text{ValidNet}(\widetilde{\gamma}), 1)$  \\ 
        $
            \widetilde{\gamma} =
            \begin{cases}
         &  (1-\varphi)\text{v} + (\varphi) \text{r} \;\;\;\,\,\;\;\; \forall \varphi \in [0, 1)  \\ 
         & (2-\varphi)\text{r} + (\varphi-1)\text{i} \;\; \forall \varphi \in [1, 2)\\
        \end{cases}$ \\
\bottomrule
    \end{tabular}
\end{wraptable}but discovers an invalid HW design. We ran this test with $\beta = 10^{x} $ for $x \in [-7, 7]$ but were not able to converge to a valid HW choice. We suspect that this is due to the irregular gradients given our sparsely valid design space. We analyze the gradient of our ValidNet model to understand this behaviour, and observe that the gradients in proximity of invalid design choices are smaller in magnitude than the gradients around valid design choices. This may mean the gradients are ineffective in navigating the design space with $L_{valid}$ given in Equation \ref{eq:update_alpha_ablation_fulydiff}.

\begin{wrapfigure}{r}{0.45\columnwidth}
\centering
\includegraphics[width=.45\columnwidth]{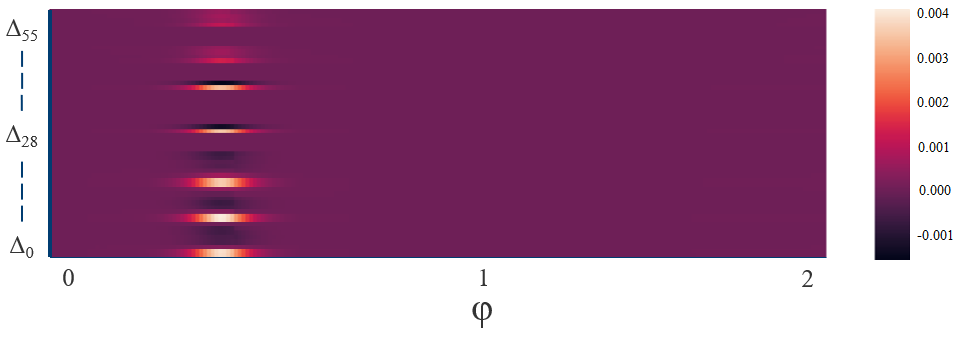}
\caption{Gradient magnitudes when interpolating from valid to invalid.}
\label{fig:interpolate_valid}
\end{wrapfigure} To study this, we measure the gradient to the HW parametrization (input to the ValidNet) with respect to a target output of 1 (where the HW parametrization is valid.). We smoothly interpolate the input HW parametrization from a valid design ($v$) to a random design ($r$) and finally to an invalid design ($i$). We record the gradient magnitude to the HW parametrization. Explicitly, we convert the information in Table \eqref{eq:hw_parametrization_validnet_update} ($\delta_{j}$) to a heat map in Figure \ref{fig:interpolate_valid}. We smoothly interpolate for $\varphi \in [0, 2)$ and use the resulting state to fetch the gradients through ValidNet.

As evident in Figure \ref{fig:interpolate_valid}, the gradients are zero almost everywhere other than the valid design point around $\varphi \in [0, 0.25]$. The observed behavior is consistent for many iterations of random sampling. As seen in our Perf-HWGEN tests previously, $\lambda \neq 0$ does not aid in the discovery of valid designs. This may be explained by the gradient study shown in Figure \ref{fig:interpolate_valid}.

\section{Conclusion}
\label{sec:conclusion}

In this paper, we demonstrate the need for an efficient solution to address issues with differentiable HW optimization (DSHWNAS) and MLP based HW generation (Exhaustive/Perf-HWGEN) on sparsely valid spaces. Our method proposes to enables co-design on sparsely valid search spaces by disentangling the training of a non-differentiable RL HW optimizer from NAS. The trained RL HW optimizer is used along with a performance predictor to guide the differentiable NAS algorithm. Our formulation is compatible wth other differentiable NAS strategies and is task agnostic (applicable to general tasks in classification, segmentation, detection, among others.). We can also adapt our RL HW optimizer to more complex HW search spaces, as explored in ConfuciuX~\cite{kao2020confuciux}. 

Sparsely valid spaces are common in HW accelerator design. Arbitrary constraints on architectural parameters can arise due to resource limitations on accelerators that support different PE (Processing Element) connectivity and compiler mapping strategies. Reducing the sparsity of valid design choices incurs a cost in hardware design/verification effort, physical design effort, fabrication cost, and energy efficiency. More importantly, it increases development time and delays deployment. We hope that the presented technique can help push accelerated deep learning into deployment faster. In future work, we hope to introduce multi-task optimization as well as develop more sample-efficient NAS methods to enable co-exploration of more complex HW search spaces.

\section{Broader Impact}
\label{sec:broader_impact}

Our work introduces a co-design method to discover efficient and realizable neural network and hardware design pairs on a broader class of hardware accelerators. Our proposed technique integrates a single-stage NAS process with an RL agent for hardware optimization. This method allows for co-design with marginally more resource requirements. For instance, training a sub-sampled architecture from our NN search space would require 160 GPU hours using the SPOSRetrain240 strategy described in the appendix. Our algorithm, RHNAS, discovers efficient NN-HW designs in only 187 GPU hours. Considering the large energy, compute, and carbon costs of previous non-differentiable NN-HW co-design works, we hope that our work will help drive more efficient NN-HW architecture search research. Our algorithm is task agnostic and can be adapted to heterogeneous workload optimization by introducing multi-workload simulation losses as a regularizer. The proposed method is particularly useful for discovering optimized configurations for ASICs before deployment. We believe that our work will encourage early co-design of accelerators in the development stages, allowing developers to make wiser decisions about the capabilities, data-flow styles, and hardware resource assignments their accelerators should have. Our modest compute requirements should allow for the deployment of accelerators in various stages of development as well as frequent prototyping of the design space. 

\bibliographystyle{nips}
\bibliography{neurips_bib}

\newpage

\appendix

\newpage
\section{Appendix}

\subsection{RL Agent Details}

\textbf{Agent details} For the policy and value networks for PPO and A2C, and Q-network for the DQN, we use MLPs with two hidden layers with sizes [64, 64].We fixed the number of training episodes to 300000 and tuned the agents learning rate, discount factor ($\gamma$) and number of steps per update across a wide range of settings.\footnote{We run all our tests in this paper on an NVIDIA DGX-2\texttrademark\; server with 4 NVIDIA V-100 32GB GPUs.}

\textbf{State representation details}:

To represent the HW embeddings, we use a flattened one-hot vector of size 36 for our HW accelerator design as there are 7 HW features we control, with two features having 2 choices, two with 4 and three with 8 (Listed in Table 1 in the paper).

As for the neural network state, we
use a concatenated representation of the one-hot choices for each layer.
 For ImageNet, we use a backbone with 20 blocks having 4 choices each - giving us a vector of length 80. For CIFAR-10 we use a smaller network with 9 blocks - giving us a vector of length 36.
 
 We utilize the same representations for the performance predictor - which we use in our reward formulation.
 We found experimentally that normalizing the rewards lead to agents learning faster - we used the VecNormalize wrapper from ~\cite{stable-baselines3}

\textbf{Hyperparameter optimization for Reinforcement Learning}\\

In order to choose the right hyperparameters, we required significant manual tuning and domain knowledge. To automate our flow and achieve further improvement, we set up hyper-parameter tuning using Optuna ~\cite{TPEBergstra, TPENIPS2011, akiba2019optuna}. Here, we focused on jointly optimizing the environment parameters ( $P$, $B$, $t_{max}$) and a few algorithm specific parameters - (learning rate, discount factor $\gamma$, number of steps per update) for all three algorithms and the number of environment steps per update for PPO and A2C. We utilized a Tree Parzen  estimator based sampler from Optuna  ~\cite{TPEBergstra, TPENIPS2011, akiba2019optuna}  and optimized each algorithm using 40 trials. We optimize agents based on how close they came to the optimal solutions on fixed test set of neural network architecture samples.

Hyper-parameter optimization significantly reduced wall time ($\approx$3x less) required to tune algorithms and enabled discovery of trained agents that achieved within 0.2 $\%$ of the optimal cycle count and 2 $\%$ of the optimal EDP values on average. 

We utilized the hyper-parameters discovered for the ImageNet RL task, and used the same settings for training our CIFAR-10 RL optimizer. We found that we were able to achieve optimality rate of 99.2$\%$ (latency) on NNs sampled for the CIFAR-10 task.

\subsection{Optimality}
In order to benchmark our RL algorithms, we compare the performance of hardware designs found by our RL solution to the optimal hardware design found through grid search using a metric Optimality $(O)$.

We measure Optimality by computing the ratio of the HW performance (Latency/EDP) of the best performing hardware on a neural network and the performance of the design suggested by the RL HW optimizer on a test set of NN architectures $(T)$. Formally, with $H_{v}$ denoting the set of all synthesisable HW designs

{ \normalsize\[O = \frac{100}{\left \| T \right \|}
 \sum_{Z_{\alpha} \in \text{T}}^{} \frac{\operatorname*{min}_{h \in H_{v}} \text{PerfPredictor}(Z_{\alpha}, h)}{ \text{PerfPredictor}(\sigma(\alpha), \text{HWOpt}(Z_{\alpha}, H_{0}))} \]
}%

Note that we only use grid search on the HW space for evaluation and for HPO (on a small subset of the NN space)
Thus, our approach achieves near optimal performance without incurring the significant cost of grid search over the whole NN space.

\subsection{Latency and EDP calculations on VTA}
We target the Intel 22FFL process~\cite{22FLTechIntel} with a 1 GHz frequency for our ASIC designs. Cycle accurate RTL simulation is used to calculate latency. For energy calculation, we assume that off-chip memory transfers take 320 pJ/Byte~\cite{Pedram_2017}. For ASIC power analysis, we ran a large VTA design synthesis for 2 million cycles at 1 GHz on our ASIC. This required 0.52W on the VTA core and 5W of power for DRAM memory transactions. Given that over 90\% of the overall power is used for DRAM access, we can leverage the data transfer count from RTL simulation (for a given neural network workload running on a particular VTA instance) and approximate energy as the product of bytes transferred and 320 pJ. 

\subsection{Predictor Network Training}
 We use L1 Loss for training the networks for 80 epochs with the Adam optimizer~\cite{kingma2017adam} with a learning rate of 0.001 and a step decay by a factor of 0.1 at epochs 40 and 60.
 
\subsection{Kendall Tau Metric}
The Kendall Tau metric ranges from -1 to 1. If $\tau = 1$, the ranking is fully preserved, if $\tau = -1$, the ranking is reversed. We measure the Kendall Tau metric by comparing the true ranking of the cycle count (as an example metric) from the test data-set for a given set of NN - HW accelerator pairs, and the ranking produced by the cycle count predictor network for the same set of NN - HW accelerator pairs.

\subsection{ImageNet search space}

\textbf{Supernet architecture    } 
\begin{table}[H]
\caption{Supernet architecture. CB - choice block. GAP - global average pooling. FC - fully connected layer. Each line describes a sequence of 1 or more identical layers, repeated \textit{Repeat} times. All layers in the same sequence have the same number of output channels. The first layer of each sequence has a stride \textit{Stride} and all others use stride 1.}\label{supernet_arch}
\centering
\begin{tabular}{c|c|c|c|c}
\hline\hline
Input & Block & Channels & Repeat  & Stride \\
\hline\hline
$224^2\times3$ & $3\times3$ Conv & 16 & 1 & 2\\
$112^2\times16$ & CB & 64 & 4 & 2\\
$56^2\times64$ & CB & 160 & 4 & 2\\
$28^2\times160$ & CB & 320 & 8 & 2\\
$14^2\times320$ & CB & 640 & 4 & 2\\
$7^2\times640$ & $1\times1$ Conv & 1024 & 1 & 1\\
$7^2\times1024$ & GAP & - & 1 & -\\
1024 & FC & 1000 & 1 & -\\
\hline\hline
\end{tabular}
\end{table}
\newpage
\subsection{CIFAR10 search space}

\textbf{Supernet architecture    } 
\begin{table}[H]
\caption{Supernet architecture. CB - choice block. GAP - global average pooling. FC - fully connected layer. Each line describes a sequence of 1 or more identical layers, repeated \textit{Repeat} times. All layers in the same sequence have the same number of output channels. The first layer of each sequence has a stride \textit{Stride} and all others use stride 1.}\label{supernet_arch_cifar}
\centering
\begin{tabular}{c|c|c|c|c}
\hline\hline
Input & Block & Channels & Repeat  & Stride \\
\hline\hline
$32^2\times3$ & $3\times3$ Conv & 64 & 1 & 1\\
$32^2\times64$ & CB & 256 & 4 & 2\\
$16^2\times256$ & CB & 640 & 4 & 2\\
$8^2\times640$ & CB & 1280 & 1 & 2\\
$4^2\times1280$ & GAP & - & 1 & -\\
1280 & FC & 10 & 1 & -\\
\hline\hline
\end{tabular}
\end{table}

\subsection{Structures of choice blocks for ImageNet and CIFAR-10}\footnote{We follow the setting including choice blocks used in the released implementation of DSNAS\cite{Hu2020DSNASDN}}    
\begin{figure}[h]
\centering
\subfloat[ Choice blocks with stride=1. Choice blocks in search space. From left to right: Choice\_3, Choice\_5, Choice\_7, Choice\_x.]{\includegraphics[width=2.5in]{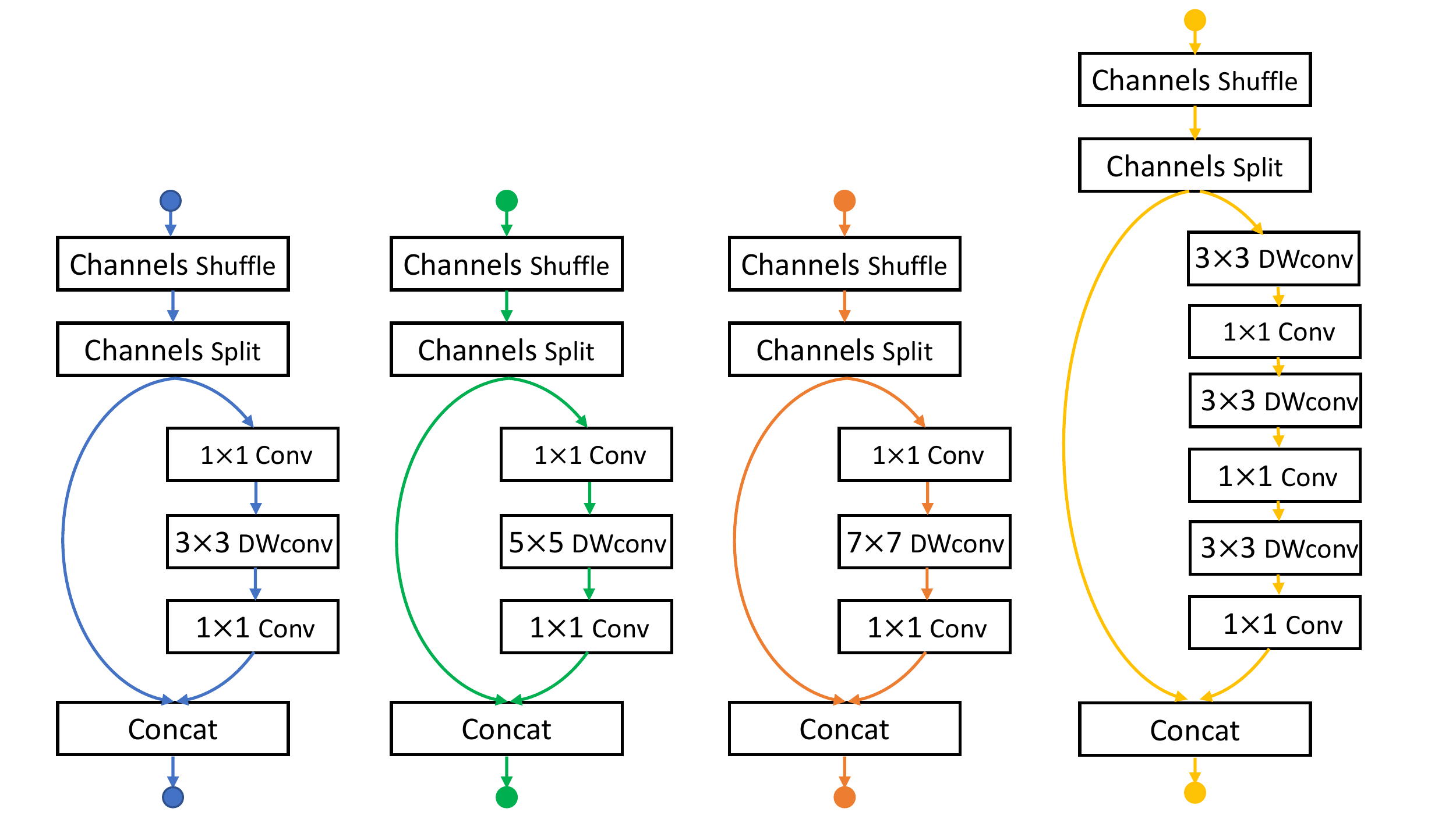}}%
\qquad
\subfloat[ Choice blocks with stride=2. Choice blocks in search space. From left to right: Choice\_3, Choice\_5, Choice\_7, Choice\_x.]{\includegraphics[width=2.5in]{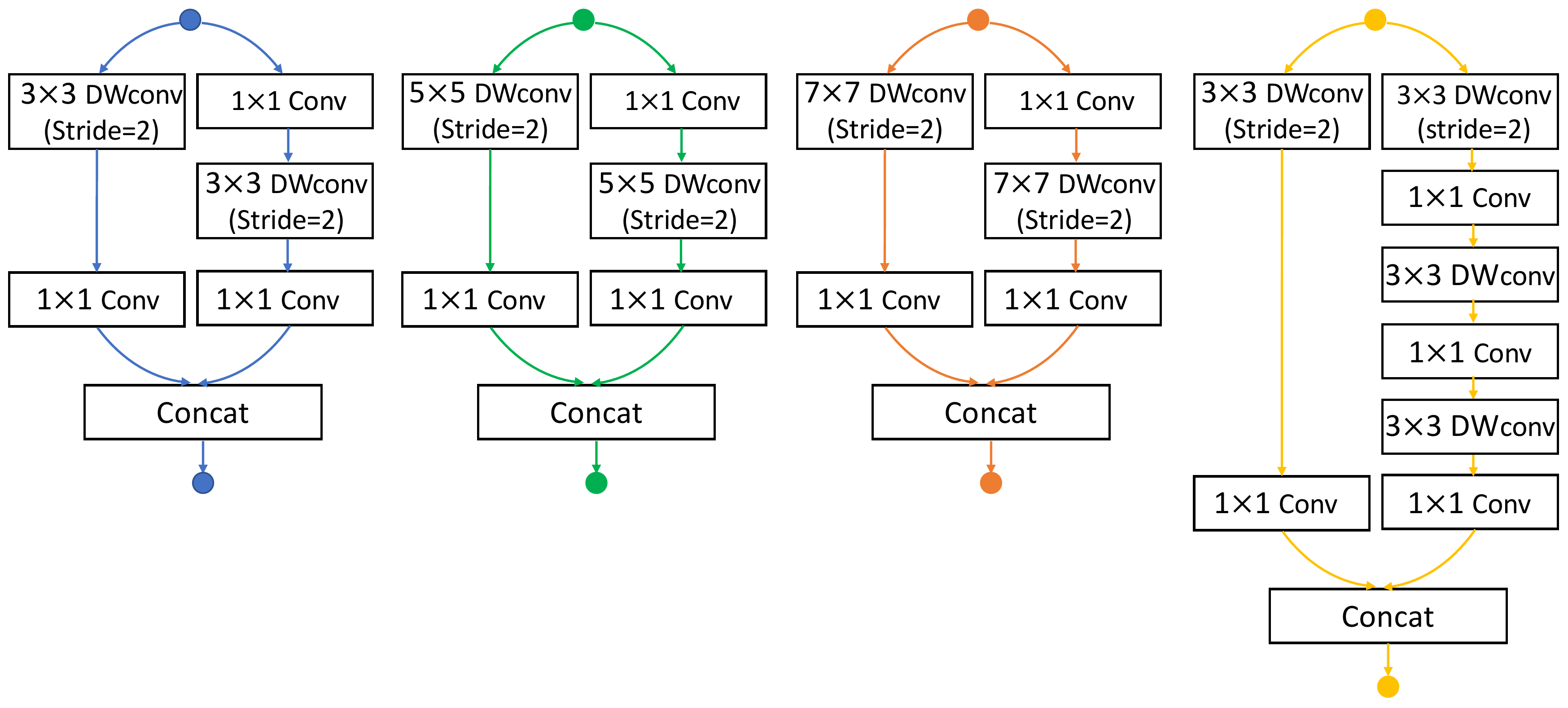}}%
\caption{Choice blocks in our search space (Figure from ~\cite{Hu2020DSNASDN})}%
\label{fig:cb_search_space}%
\end{figure}
\subsection{CIFAR-10 experimental details}
We first do 80 epochs of NN architecture search. The NN architecture is then trained from scratch using the SPOSretrain240 strategy. We found that this gave better accuracy for CIFAR-10. Thus, the total number of epochs required for CIFAR-10 search and train is 320. \\
\textbf{SPOSretrain240}\\
Epochs: 240 \\
Batch size: 1024 \\
Input data has been augmented with random crops, random horizontal flops.  \\
Torch Transforms Normalize: transforms.Normalize((0.4914, 0.4822, 0.4465), (0.2023, 0.1994, 0.2010)) \\
Categorical Cross Entropy \\
SGD: Momentum = 0.9, weight decay=0.00004 \\
lr\_mode : cosine \\
base\_lr: 0.34 \\
targetlr: 0.0 \\
power: 1 \\
warmup\_mode: linear \\
warmup\_lr: 0.0001 \\
warmup\_epochs: 5 \\

\subsection{ValidNet MLP}
This MLP has two hidden layers with ReLU activation function and a SoftMax output to indicate whether an input HW accelerator design is valid. The data-set is generated by randomly sampling design choices and their output from our analytical validity checker. Since there are very sparsely distributed valid HW design choices, we train the Validity MLP network with a cross entropy loss and a weighted loss vector $[0.05, 0.95]$ for valid and invalid designs respectively. The network is trained for 150 epochs and samples 10240 design points. This validity checker MLP delivered an accuracy of $>99\%$.

\subsection{Performance Predictor Interpolation test. }

We utilize the $\sigma(\alpha)$ NN categorical distribution in RHNAS. Thus we need to ensure that our performance predictor is able to take in probability distribution of neural network architectural choices and estimate the expected performance.

 To do this, we conduct a test where we compute the ratio between the performance predictor output for a continuous NN categorical distribution to the expected value for this output (Equation \ref{eq:interpolation_test_perfpred_left}). We obtain this expected value by summing over the predicted metric from discrete architectures sampled from the categorical distribution obtained from $\sigma(\alpha)$ multiplied by their likelihood of being sampled. We found that the ratio is $0.9910$ with a variance of $0.0012$ after sampling over 400 NN distributions. This indicates that the performance predictor is able to accurately interpolate performance between NN architectural choices.

\begingroup
\allowdisplaybreaks
\begin{align} 
    \begin{split}
    & \frac{\text{PerfPredictor}(\sigma(\alpha), H_{0})}{\displaystyle \mathop{\mathbb{E}}_{Z_{\alpha} \sim 
 \sigma(\alpha)}(\text{PerfPredictor}(Z_{\alpha}, H_{0}))} 
  \vspace{-0.1cm}\label{eq:interpolation_test_perfpred_left}
    \end{split}
\end{align}
\endgroup

\end{document}